\documentclass{article}

\usepackage{natbib}
\usepackage{arxiv}

\usepackage[utf8]{inputenc} 
\usepackage[T1]{fontenc}    
\usepackage{url}            
\usepackage{booktabs}       
\usepackage{amsfonts}       
\usepackage{nicefrac}       
\usepackage{microtype}      
\usepackage{lipsum}		
\usepackage{graphicx}
\usepackage{doi}

\usepackage{amsmath}
\usepackage{tikz}
\usepackage{dutchcal}
\usepackage[caption=false,font=footnotesize]{subfig}

\title{Learning What to Memorize: Using Intrinsic Motivation to Form Useful Memory in Partially Observable Reinforcement Learning}

\author{ \href{https://orcid.org/0000-0003-2646-4850}{\hspace{1mm}Alper Demir}\\
	Department of Computer Engineering\\
	Izmir University of Economics\\
	Izmir, Turkey \\
	\texttt{alper.demir@ieu.edu.tr} \\
}

\date{}

\renewcommand{\shorttitle}{Learning What to Memorize}

\hypersetup{
pdftitle={A template for the arxiv style},
pdfsubject={q-bio.NC, q-bio.QM},
pdfauthor={David S.~Hippocampus, Elias D.~Striatum},
pdfkeywords={First keyword, Second keyword, More},
}

\begin{document}
\maketitle

\begin{abstract}
Reinforcement Learning faces an important challenge in partial observable environments that has long-term dependencies. In order to learn in an ambiguous environment, an agent has to keep previous perceptions in a memory. Earlier memory based approaches use a fixed method to determine what to keep in the memory, which limits them to certain problems. In this study, we follow the idea of giving the control of the memory to the agent by allowing it to have memory-changing actions. This learning mechanism is supported by an intrinsic motivation to memorize rare observations that can help the agent to disambiguate its state in the environment. Our approach is experimented and analyzed on several partial observable tasks with long-term dependencies and compared with other memory based methods. \footnote{Our code is available at \href{https://github.com/gliese876b/smm}{https://github.com/gliese876b/smm}}
\end{abstract}


\section{Introduction}
Reinforcement Learning (RL) has proved itself as a feasible machine learning method in problems that involve sequential decision making \citep{sutton_reinforcement_1998}. Many algorithms guarantee to converge to an optimal policy when the state of the agent is clearly apparent. However, such optimality guarantee vanishes under partial observability where the agent is no longer capable of receiving necessary information about the task to be learned. In such environments, the observation space may be smaller than the state space and be ambiguous so that an observation may correspond to different states calling for different optimal actions. In order to overcome this issue, the agent may benefit from an additional memory, storing previous observations. This way, it can clarify its state to learn a policy solving the task.

Besides, finding a good policy for a problem, the speed of learning is another focus of Reinforcement Learning. A relatively alternative approach to improve RL is intrinsic motivation where the reward function is supported by an additional intrinsic reward \citep{Singh2010}. This reward is determined by a motivation that is persistent throughout the agent's life and is driven by aspects like curiosity, novelty and surprise. Applying intrinsic motivation enables the agent to be adaptable to various problems and explore them better, which is more consistent with the idea of lifelong learning. Intrinsic motivation is coupled by many RL studies, proving its advantages \citep{Bellemare2016,zhang2019scheduled}. However, this concept is not yet introduced to memory management under partial observability.

Many algorithms that tackle RL under partial observability, devised a fixed method that manages the memory. Such methods include keeping a fixed or variable length window of information gathered previously in the external memory \citep{lin1992memory,mccallum_reinforcement_1996}, learning a finite state machine \citep{Meuleau1999}, predicting state representations \citep{littman2002predictive} and learning through long short-term memory \citep{Zhu2018,mnih2016asynchronous,wang2016sample}. However, these methods are limited to certain problems because of their strict assumptions about them. 
Our argument is that the control of the memory can be given to the agent and the agent can \textit{learn} what to store in it based on the task \citep{Peshkin1999}. Such an idea allows the agent to be more adaptable for different tasks, especially for the ones that possess long-term dependency between an observation and the corresponding action. In these problems, the observation that affects the optimal action at a time step is taken some unfixed time steps ago, so the content of the external memory should be changed based on the dynamics of the problem. Rather than directly following a fixed method to determine the memory content, the agent can be intrinsically rewarded for keeping observations in the memory, which will be useful in the future. This approach can motivate the agent to form an external memory with rare perceptions without ignoring the dynamics of the problem.

In this paper, we introduce the notion of intrinsic motivation to memory based RL under partial observability. We enhance the action set of the agent to include actions that can modify the memory and provide an intrinsic reward based on the content of the memory that the agent decides to keep. Our work is inspired by a previous study \citep{Peshkin1999}, yet we form the memory in a different way and also, support this memory management learning process by intrinsic motivation to keep rare observations. Our experiments include tests on partially observable tasks that require long-term dependency and comparisons with the original approach and other similar methods.

\section{Background and Related Work}
In this section, we provide a brief summary of the topics that our study is based on. The section defines the environment model for partial observability, mentions several reinforcement learning approaches to address the problem, and discusses a concept, namely intrinsic motivation, to improve learning performance together with its implementations in RL. 

\subsection{POMDPs}
A general model for realistic problems with limited information is given by the Partially Observable Markov Decision Process (POMDP). A POMDP is defined by a tuple $\left\langle S, A, T, R, \Omega, O\right\rangle $ where $S$ is a finite set of states, $A$ is a finite set of actions, $T : S\times A \times S \rightarrow [0, 1]$ is a transition function, $R : S \times A \times S \rightarrow \Re$ is a reward function, $\Omega$ is a finite set of observations, and $O:S \times A \rightarrow \Pi(\Omega)$ is the observation function \citep{kaelbling_reinforcement_1996}.

The POMDP model has been addressed in two ways in the literature. The first of these aims to keep a \textit{belief state} \citep{astrom1965optimal}, which is a probability distribution over the set of states, by assuming that the agent has access to the state space of the problem and the transition function, and can update this belief state with observations from the environment. The other is the \textit{hidden state POMDP approach} \citep{mccallum_reinforcement_1996}, which assumes that the agent has no access to the problem structure and only takes observations from the environment. Under this setting, the agent needs to keep an \textit{estimated state} to base its actions upon. In this study, we follow the latter approach.

\subsection{Reinforcement Learning in POMDPs with Hidden States}
Many reinforcement learning methods that have proven themselves in fully observable environments, fail in partially observable ones and lose the guarantees of reaching the optimal policy \citep{chrisman_reinforcement_1992}. The main reason for this is the \textit{perceptual aliasing} problem that occurs in such environments \citep{whitehead_learning_1991}. Perceptual aliasing happens when more than one state corresponds to the same observation. In this case, the agent cannot distinguish between different states, where the best action may be different, and therefore cannot find the optimal policy. This situation makes the task of reinforcement learning in POMDPs with hidden states a difficult and unexplored field.

Methods aimed at RL in such environments are divided into two; memory-based and memoryless. Memoryless methods aim to converge to the best policy without an additional storage of the information gathered from the environment. In such problems, finding the best memory-free policy is classified as NP-Hard \citep{littman_memoryless_1994}. On the other hand, the eligibility trace approach has been shown to be useful in practice. One of the methods using eligibility traces, Sarsa($\lambda$), is an on-policy learning algorithm that updates all of the entries in the value function at a step based on their eligibility \citep{loch_using_1998}.

Memory based approaches depend on the addition of memory to tackle the problem of partial observability. In an environment with limited sensory information, the agent has to predict the state it is in, based on the observations it receives for the goal of finding the right course of action. To avoid perceptual aliasing, the agent must use a memory where it can hold past observations, actions, or rewards. Using this memory, the agent forms an \textit{estimated state} for its place in the state space.

The simplest method of generating state estimation using memory is to keep a fixed length window \citep{lin1992memory}. In this method, the current state estimation is formed by adding the current observation to the external memory where a fixed number of past observations are kept. The agent acts based on this state estimation to solve the given task. A better approach is to increase the memory as needed and allow it to have different lengths. Utile Suffix Memory, Nearest Sequence Memory, and U-Tree methods are examples of such variable length memory methods \citep{mccallum_reinforcement_1996,Zheng2011}. For example; USM and U-Tree form a tree of observations and actions where each leaf node contains a set of values for each action. Each branch in this tree can be considered as an estimated state which the agent forms its policy on.

Unlike window based methods, Long Short-Term Memory approach, which uses deep neural networks, has proven itself in partial observable problems based on visual data \citep{hochreiter1997long,Bakker2001,Le2018,Zhu2018}. In this approach, there are forget gates in the recurrent neural network in addition to input and output gates, so that this internal memory is not constrained by a time interval and can hold useful information for a long-term.

One study that holds external memory is keeping it as a finite state machine and generating it online by using stochastic gradient descent \citep{Meuleau1999}. Such a finite state machine is capable of representing every state of an environment by storing all gathered information. Another study uses past observations of the agent to generate predictive state representations based on future observations and action predictions \citep{littman2002predictive,Singh2003,James2004}. The study of \citep{ToroIcarte2019} proposes an approach where an external reward machine is learned with the help of a labelling function in order to represent the reward function and the problem is decomposed into smaller solvable tasks.

The study inspired this paper is aimed to keep an external memory as a binary string and let the agent learn its management. In this way, the memory is not controlled by a fixed method but can be managed by the agent according to the needs of the problem \citep{Peshkin1999,Lanzi2000}. VAPS(1) algorithm is used to learn a policy by stochastic gradient descent, where the agent has additional actions to set the memory to any value at a time. VAPS(1) is shown to learn a good policy in POMDPs with long-term dependencies.

Methods that use the window-based memory are based on the assumption that the required distinctive information is temporally close to the current moment, but this assumption is not always realistic. The LSTM approach, which is independent of the time interval, also faces the problem of the neural network being too customized for each environment. Although many studies in the literature present deep networks that show good RL performance in an environment, these deep networks may be too specialized in a level that cannot show learning in a different problem and may be limited to problems based on visual data \citep{Henderson2018}. Then, the predictive state representation approach requires a comprehensive exploration of the infrastructure of the problem to achieve good results and the reward machines require an additional labelling function to be formed during learning. Finally, in the approach where external memory is kept as a binary string, the memory is unrelated from the observations and the order of gathering them. There is a room for improvement for the task of learning memory management in RL, which is an undiscovered field.

\subsection{Intrinsic Motivation}
Intrinsic motivation, inspired by the field of psychology, has been used in Reinforcement Learning. According to this concept, the agent receives intrinsic rewards from the satisfaction of the action itself, independent of an external goal. Therefore, this motivation never disappears and continues every time the internal reward is active. Examples of concepts that cause intrinsic motivation are curiosity, novelty, and surprise \citep{Singh2010}.

In the traditional reinforcement learning structure, the rewards are determined by the model and given outside of the learning mechanism. Recent studies show that, in addition to external reward, intrinsic rewards contribute to learning performance of the agent \citep{Singh2010,barto2013intrinsic}. The new reward function which determines the rewards the agent receives, is defined as $\hat{R}(s, a, s') = R(s, a, s') + R^{int}(s, a, s')$ where $R^{int}(s, a, s')$ represents the intrinsic motivation that corresponds to reaching state $s'$ by taking action $a$ at the state $s$.

The use of intrinsic motivation in reinforcement learning can be addressed under two main headings \citep{Aubret2019}. The first of these is gaining knowledge. In this approach, solutions are sought for the problems of exploration and state representation in RL by giving intrinsic rewards to the actions that enable the agent to get more information from the problem. Studies show that the intrinsic rewards given through perception novelty enable the agent to explore and learn the problem better \citep{Bellemare2016,ostrovski2017count}. 

Another use of intrinsic motivation is the problem of skill learning. The aim here is to learn different and generalizable skills by using intrinsic motivation or to infer which skills should be learned by following a curriculum. With this approach, it has been shown that with predefined intrinsic rewards, learning performance improves in problems requiring hierarchy where external reward is sparse \citep{kulkarni2016hierarchical,zhang2019scheduled}.

Although the concept of intrinsic motivation has been studied extensively in fully observable reinforcement learning problems, the number of studies in partially observable problems is quite limited. This situation arises from the difficulty of learning through the lack of necessary information. Hence, the effect of intrinsic motivation in partially observable environments remains an unexplored area.

\section{Self Memory Management}
Partial observability creates a significant challenge for Reinforcement Learning; having limited and aliased information about the environment, the agent is unable to devise a policy that can perform well. In order to overcome this issue, it is required to keep a memory of the history, besides the current observation, so that the different states that map to the same observation can be distinguished. 

Our study follows the concept of keeping a memory under partial observability and is inspired by the approach that gives the control of the memory to the agent \citep{Peshkin1999}. Unlike several studies which use a fixed method to control the memory, our approach, that we name as Self Memory Management (SMM), puts the agent in charge and allow it to modify the memory by augmenting the action space with memory-changing actions. This learning processes is supported by providing an intrinsic motivation to keep key information in memory. In an ambiguous environment, relatively rare observations become distinctive and can help the agent to disambiguate its state. An intrinsic reward can be consistently provided based on the memory content, so that memories with uncommon observations are supported.

Such an approach possesses several advantages. First, it enables the agent to be more flexible according to the dynamics of the problem. Each environment may require a different memory content based on the task to be learned. An agent may need to remember an old observation to make the best decision in future in one problem whereas it may need to remember just the previous one to get a sense of direction in the other. With SMM, the agent can learn the best memory content which allows it to solve the task, while interacting with the environment. Second, the idea of keeping rare observations in memory is not forced but supported. Thus, what the task requires is not ignored and added into agent's decision making. The agent can learn to store a rare observation in memory when it is useful for the task.

Following sections defines the form of the memory, which is different from the one in the study of Peshkin et al \citep{Peshkin1999}, the intrinsic reward that our approach provides and the overall algorithm.

\subsection{Defining Memory and Actions over Memory}
Our definition of a \textit{memory} $m_t$ at time $t$ is a sequence of observations that is previously perceived and is notated as;
\begin{align*}
    m_t = \langle o_i, o_j, ..., o_k \rangle
\end{align*}
where $o_i, o_j, ..., o_k$ represent the observations at times $i, j, k$ respectively and $i < j < k < t$. That is, the observations are ordered from left to right according to the time that they are gathered. Note that $i, j, k$ and $t$ are not necessarily consecutive numbers. An empty memory $m = \langle \rangle$ has the size of zero, that is $|m| = 0$.

Our method composes the action space with a set of actions modifying the memory. The set of memory actions $\mathcal{A}$, that we have in this study, consists of two actions; $\mathcal{A} = \{ \mathcal{a}^{push}, \mathcal{a}^{skip} \}$ and the effects of these actions are defined by a function $\mathcal{T}: M \times \Omega \times \mathcal{A} \rightarrow M$ where $M$ defines the set of all possible memories. 

At time $t$, for a memory $m_t = \langle o_i, o_j, ..., o_k \rangle$ and the current observation $o_{t}$, a memory action $\mathcal{a}_t$ determines the next memory $m_{t+1}$ according to
\begin{align*}
     m_{t+1} &= \mathcal{T}(m_t, o_{t}, \mathcal{a}_t) = 
                    \begin{cases}
                        \langle o_i, o_j, ..., o_k \rangle, & \text{if } \mathcal{a}_t = \mathcal{a}^{skip},\\
                        \langle o_i, o_j, ..., o_k, o_{t} \rangle, & \text{if } \mathcal{a}_t = \mathcal{a}^{push} \text{ and } |m_t| < c, \\
                        \langle o_j, ..., o_k, o_{t} \rangle, & \text{if } \mathcal{a}_t = \mathcal{a}^{push} \text{ and } |m_t| = c, \\
                    \end{cases}
\end{align*}
where $c \geq 0$ is the capacity of the memory, a parameter to bound it. With this definition, taking the action $\mathcal{a}^{push}$ pushes $o_{t}$ to the end and if the memory is at its full capacity, pops the oldest observation $o_i$ from it. On the other hand, the action $\mathcal{a}^{skip}$ skips $o_{t}$ without changing the memory.

The advantages of defining the memory in this way is two-fold. First, keeping the observations in a sequence enables the memory to keep them ordered, giving the agent a sense of direction. By this definition, a memory $\langle o', o'' \rangle$ is a different memory than $\langle o'', o' \rangle$, hence it represents a different experience of the problem. Second, $\mathcal{T}$ restricts the transitions between memories. The agent cannot jump between completely distinct memories, therefore the set of visited memories is kept bounded according to the observations gathered from the environment.

\subsection{Using Intrinsic Motivation}
The memory management method proposed in this study is based on learning what to keep in the memory. This learning process can be supported by giving intrinsic motivation to changes on memory. Our method attacks the problems where there is long-term dependency between the information and the corresponding action. That is, the required observation to take an optimal action at a moment is temporally distant. This kind of problems require the agent to keep the observation in memory for the time that it is going to be important. 

We argue that such observations are rare in an ambiguous environment. Therefore, an agent motivated to keep rare observations in memory can improve its learning performance. In this paper, we define the intrinsic motivation to keep a memory $m$ as;
\begin{align*}
    R^{int}(m) = \beta \cdot \left[ \left( \sum_{o \in m} (1 - P(o)) \right) - c \right]
\end{align*}
where $c \geq 0$ is the memory capacity, $\beta \in [0, 1]$ is a parameter to control the intrinsic reward and $P(o) \in [0, 1]$ is the probability of gathering observation $o$. This probability can be calculated by keeping a count of each visit of an observation and forming a probability distribution over observations based on their frequency of occurrence. 

Note that $R^{int}(m)$ is bounded to $[-c, 0]$ but the magnitute of the reward decreases with the addition of infrequent observations to the memory. Such a function motivates the agent to utilize the memory to its full capacity by keeping rare observations in it. Also, by being a negative value, it does not diverge the agent away from a goal state, keeping the main aim of it unchanged.

\subsection{Overall Method}
Having an external memory, the agent needs to form a \textit{state estimation} by using it in order to make better decisions. 

At a time $t$, the agent's state estimation $x_t$ is formed by concatenating the memory $m_t = \langle o_i, o_j, ..., o_k \rangle$ with the current observation $o_t$ as below;
\begin{align*}
    x_t = m_t + o_t = \langle o_i, o_j, ..., o_k, o_t \rangle
\end{align*}
where $o_i, o_j$ and $o_k$ are previously seen observations at times $i, j$ and $k$ in the memory. The set of estimated states, formed by this approach, is notated as $X$.

A composed set of actions $\hat{A} = A \times \mathcal{A}$ is formed by using the environment actions $A$ and the memory actions $\mathcal{A}$. With these definitions, the agent aims to learn a policy $\pi: X \times \hat{A} \rightarrow [0, 1]$. 

At a time $t$, with the estimated state $x_t$ formed by the memory $m_t$ and the current observation $o_t$, the agent takes a composed action $(a_t, \mathcal{a}_t)$ according to its current policy $\pi_t$ where $a_t$ is the action to pass to the environment and $\mathcal{a}_t$ is the action over the memory. Taking action $a_t$ in the environment results in the next observation $o_{t+1}$ and the agent's next memory $m_{t+1}$ is determined according to $\mathcal{T}(m_t, o_t, \mathcal{a}_t)$. The next estimated state $x_{t+1}$ is formed by concatenating $m_{t+1}$ with $o_{t+1}$. The agent gets a reward $\hat{r}_t$ according to the new reward function;
\begin{align*}
    \hat{r}_t &= \hat{R}(x_t, (a_t, \mathcal{a}_t), x_{t+1}) = R(x_t, (a_t, \mathcal{a}_t), x_{t+1}) + R^{int}(x_t, (a_t, \mathcal{a}_t), x_{t+1})
\end{align*}
where $R(x_t, (a_t, \mathcal{a}_t), x_{t+1})$ reduces to the original reward function $R(s_t, a_t, s_{t+1})$ and $R^{int}(x_t, (a_t, \mathcal{a}_t), x_{t+1}) = R^{int}(m_{t+1})$ as the intrinsic motivation to keep the memory $m_{t+1}$.

These modifications alters the setting that the learning mechanism operate on while the interaction between the agent and the environment is kept the same. The agent provides an action from $A$ and gets observations and rewards from $\Omega$ and $R$. Moreover, the proposed method is independent of the used reinforcement learning algorithm. Different learning methods can be used to learn the agent's policy.

We implemented our approach on top of Sarsa($\lambda$) algorithm \citep{loch_using_1998} that shows faster convergence under partial observability. In this setting; we have a Q value for each $(x, (a, \mathcal{a}))$ pair and at a time $t$, the agent experiences $\langle x_t, (a_t, \mathcal{a}_t), \hat{r}_t, x_{t+1} \rangle$ and follows the update rules;
\begin{equation}
\begin{aligned}
\eta_{t}(x_{t}, (a_t, \mathcal{a}_t)) = 1
\end{aligned}
\end{equation}
\begin{equation}
\begin{aligned}
\forall(x \neq x_{t} &\text{ or } (a, \mathcal{a}) \neq (a_t, \mathcal{a}_t)), \; \eta_{t}(x, (a, \mathcal{a})) = \gamma  \lambda  \eta_{t-1}(x, (a, \mathcal{a}))\\
\end{aligned}
\end{equation}
\begin{equation}
\begin{aligned}
&\forall x, (a, \mathcal{a}), \; Q_{t+1}(x, (a, \mathcal{a})) = Q_{t}(x, (a, \mathcal{a})) + \alpha  \eta_{t}(x, (a, \mathcal{a}))  \delta_{t}
\end{aligned}
\end{equation}
where $\eta_{t}(x, (a, \mathcal{a}))$ is the eligibility trace of the estimated state - composed action pair $(x,(a, \mathcal{a}) )$, $\alpha$ is the learning rate, $\lambda$ is the trace decay constant, $\gamma$ is the discount factor and $\delta_{t} = \hat{r}_{t} + \gamma Q_{t}(x_{t+1}, (a_{t+1}, \mathcal{a}_{t+1})) - Q_{t}(x_{t}, (a_t, \mathcal{a}_t))$ is the TD-error at time step $t$.

\section{Experiments}
We experimented on three partially observable environments that can show the effect of our approach. Each of these have a long-term dependency, requiring the agent to keep a temporally distant observation in memory to the point where it is useful. 
We picked Sarsa($\lambda$) as the underlying learning algorithm for our approach. We compared Sarsa($\lambda$) with self memory management (abbreviated as Sarsa($\lambda$) w/ SMM), with Sarsa($\lambda$) with a memory of fixed window (abbreviated as Sarsa($\lambda$) w/ FW), and the incremental implementation of VAPS(1) algorithm \citep{Peshkin1999} since it is the original study proposing the idea of agent controlled memory. 
VAPS(1) algorithm is implemented differently in terms of the action space: In the original study, the authors add a new action for setting each bit of the memory but taking this action wastes a time step making no changes in the environment. In order to have a fair comparison, we implemented their second approach, which allows the agent to have composite actions that can directly set the entire memory to a value. 

In addition to methods with external memory, the experiments consist of two baselines that use LSTM to summarize the history of the agent; A3C \citep{mnih2016asynchronous} and ACER \citep{wang2016sample}. 

\subsection{Setup}
The experiments are presented in terms of two aspects; learning performance and the number of times the external memory is changed, that is $m_t \neq m_{t+1}$. These metrics are selected so that our approach can be shown to improve the learning speed by learning how to manage the memory with minimal changes.

Both of the Sarsa($\lambda$) versions are tested with $\epsilon$-greedy action selection with $\epsilon$ decaying from $0.2$ to $0.001$ and the rest of the parameters are set as $\lambda = 0.9$, $\alpha = 0.01$, $\gamma = 0.9$ whereas Sarsa($\lambda$) with SMM used $\beta = 1.0$ in the main experiments. VAPS(1) algorithm follows Boltzmann law to select actions, having a stochastic policy. The Boltzmann temperature value is decayed over episodes, as the original study suggests.

For A3C and ACER, we used Stable Baselines implementations \citep{stable-baselines}. We followed a setting similar to the study of \citep{ToroIcarte2019}. Both of the methods used a network of 5 fully connected layers with 64 neurons per layer. During learning, the methods used 32 sampled experiences from a buffer of size 100,000 with the learning rate of $0.01$. We selected the number of hidden neurons for the LSTM as 128, where the rest of the parameters are left as default.

Due to the stochastic process in each algorithm, we present the average results over 50 experiments where each episode either ended when the agent reached to the goal state or to the step limit of 500.

\subsection{\texttt{Load/Unload}}

\begin{figure}[h!]
    \centering
    \includegraphics[scale=0.6]{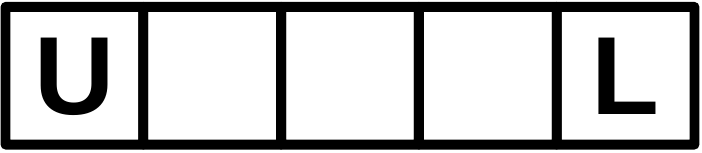}
    \caption{Sketch of the \texttt{Load/Unload} problem where the locations marked with $U$ and $L$ determine the states that the agent gets unloaded and loaded respectively. The observations are limited to the presence of a wall in four navigational directions.}
    \label{fig:load_unload}
\end{figure}

\texttt{Load/Unload} (Figure \ref{fig:load_unload}), is a simple problem \citep{Peshkin1999} where the agent is capable of loading at location $L$ and unloading at location $U$. An episode starts with the agent unloaded at $U$ and ends when the agent reaches the location $U$ after getting loaded at $L$. The agent has two deterministic actions to go \textit{west} and \textit{east} and is rewarded with 1.0 only when it reaches the location $U$ as loaded. The problem is partially observable since the agent only gets observations based on a presence of a wall in four directions, but it does not know whether it is loaded or not. The agent has to keep a memory of being loaded, on the states in the corridor where each of them provides the same observation. Overall, the domain has $|S| = 8, |A| = 2, |\Omega| = 3$. This is the environment which VAPS(1) is experimented on, with a 1 bit of memory in the original study.

\begin{figure}[h!]
	\centering
	\subfloat[]{
	    \centering
		\includegraphics[scale=0.2]{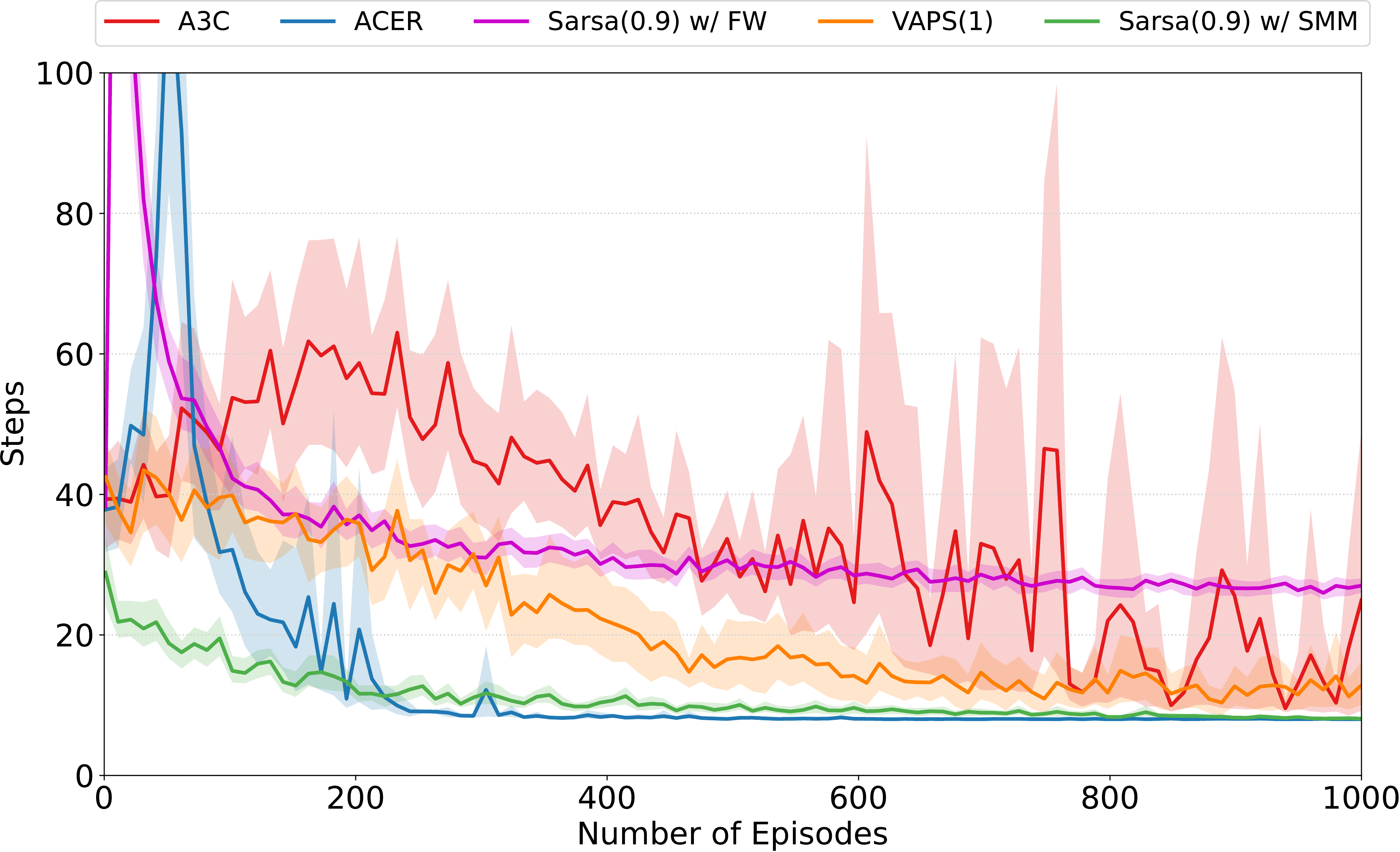}
        \label{fig:steps_load_unload}
	} 
	\subfloat[]{
	    \centering
		\includegraphics[scale=0.2]{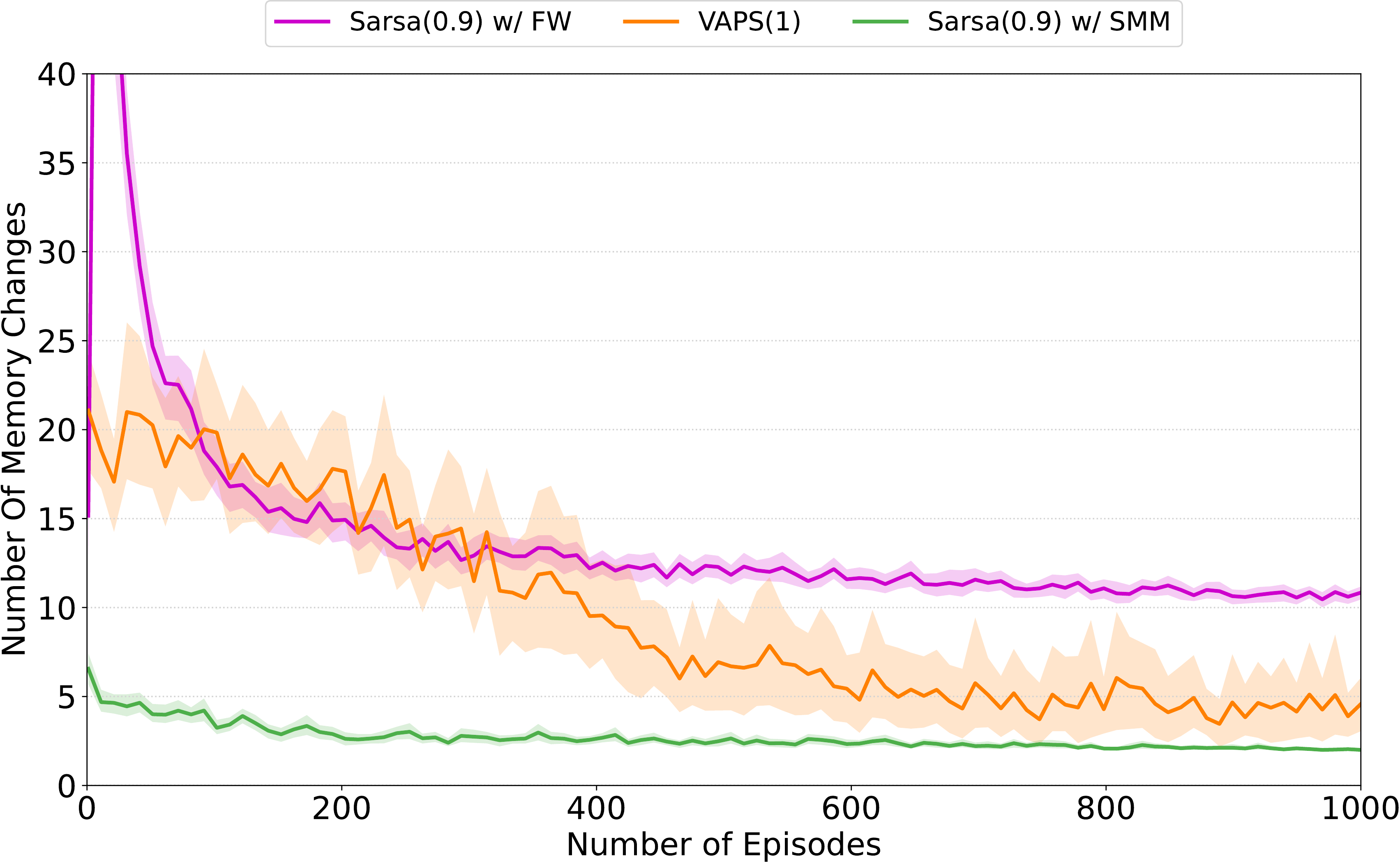}
		\label{fig:memory_changes_load_unload}
	}
	\caption{(a) Number of steps to complete an episode, (b) number of changes on external memory during an episode in \texttt{Load/Unload} problem. VAPS(1) algorithm has a 1 bit memory where both Sarsa($\lambda$) with FW and Sarsa($\lambda$) with SMM has a memory with a capacity of 1 observation. The lines represent the average of 50 experiments and shaded areas are the $95\%$ bootstrapped confidence intervals.}
	\label{fig:result_load_unload}
\end{figure}

In order to demonstrate the speed of learning, Figure \ref{fig:steps_load_unload} shows the number of steps to complete an episode in the problem, rather than total collected reward, since the environment provides a reward only when the agent reaches to the goal state. It shows that a fixed window memory fails to compete with the other methods that show convergence to a better policy. Sarsa($\lambda$) with SMM, starts strong on reaching to the goal state in lesser number of steps. Since the observations at the loading and unloading stations are relatively rare, the agent with SMM is motivated to keep them in memory, leading to a better learning performance. While VAPS(1) agent has to learn when to set or clear its memory, SMM is supported by intrinsic motivation, thus it learns what to store in the memory much quickly. Both of the LSTM-based methods show convergence, but they are less sample-efficient compared to SMM. Also, it can be seen in Figure \ref{fig:memory_changes_load_unload}, this learning performance of SMM is achieved with a minimum number of changes over the memory, compared to the other external memory based methods. In fact, SMM updates the memory only when the agent reaches to an end of the corridor but leaves it unchanged in the middle part.

\subsection{\texttt{Modified Meuleau's Maze}}

\begin{figure}[h!]
    \centering
    \includegraphics[scale=0.3]{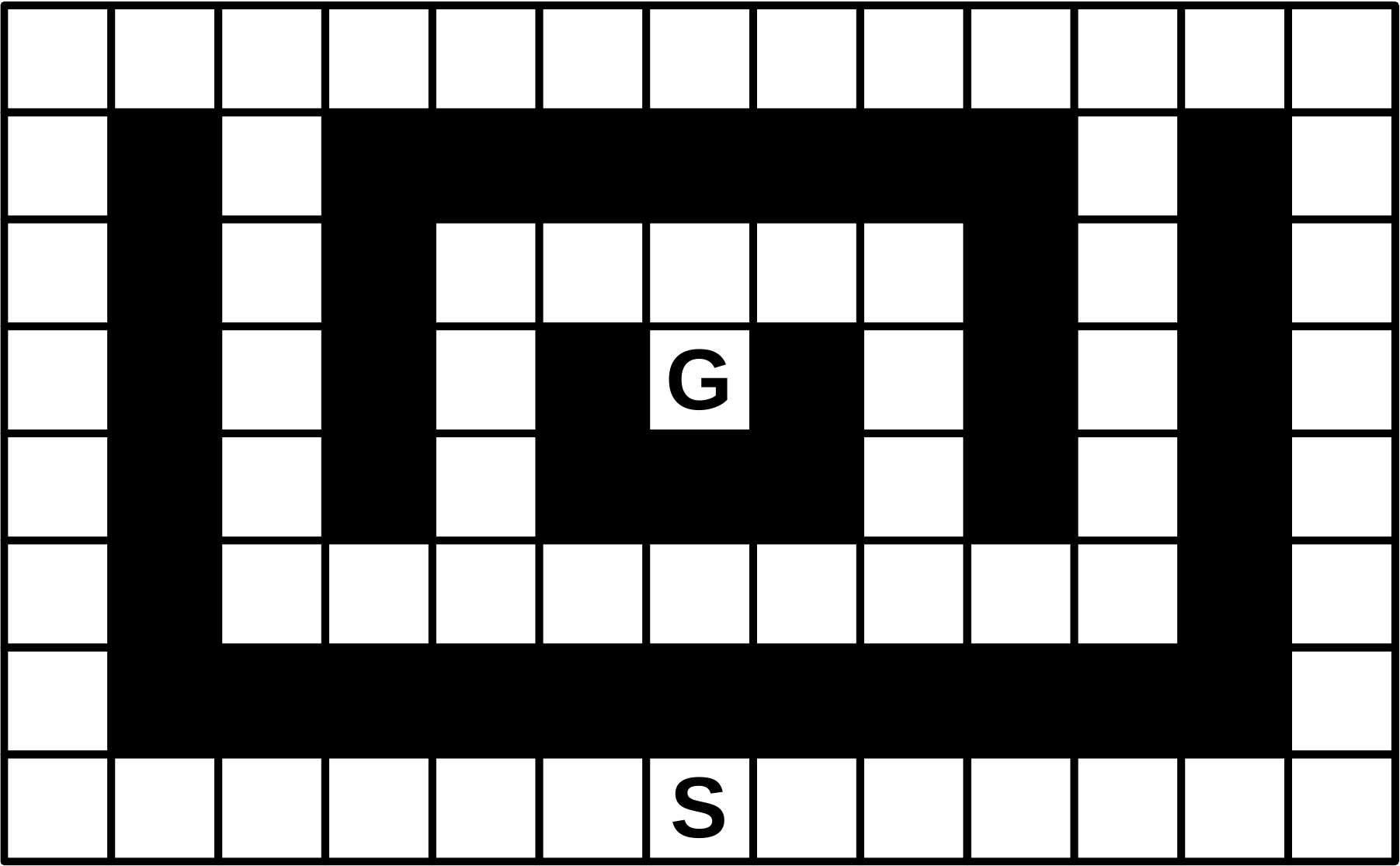}
    \caption{Sketch of \texttt{Modified Meuleau's Maze} where the start and the goal states are marked with $S$ and $G$, respectively. The observations are limited to the presence of a wall in four navigational directions.}
    \label{fig:meuleau_maze}
\end{figure}

Another partially observable environment we tested on is \texttt{Modified Meuleau's Maze} \citep{Meuleau1999S}, shown in Figure \ref{fig:meuleau_maze}. The agent has four navigational actions as \textit{north}, \textit{east}, \textit{south}, \textit{west}, starts from the state with $S$ and aims to reach the goal state with $G$. The observations are partial in terms of the presence of a wall in four directions and the agent is rewarded for reaching $G$ with 5.0 where other actions get -0.01. The actions are stochastic in this version, where the agent is capable of moving to the intended direction with 0.8 probability and can end up in a random direction with 0.2 probability. In this problem, the states in the corridors provide aliased observations and may require opposite actions (like in the vertical corridors) and unlike other domains, the agent has to update its memory whenever it reaches to a corner or a T-junction, rather than keeping a fixed memory content. The environment has $|S| = 65, |A| = 4, |\Omega| = 8$.

\begin{figure}[h!]
	\centering
	\subfloat[]{
	    \centering
		\includegraphics[scale=0.2]{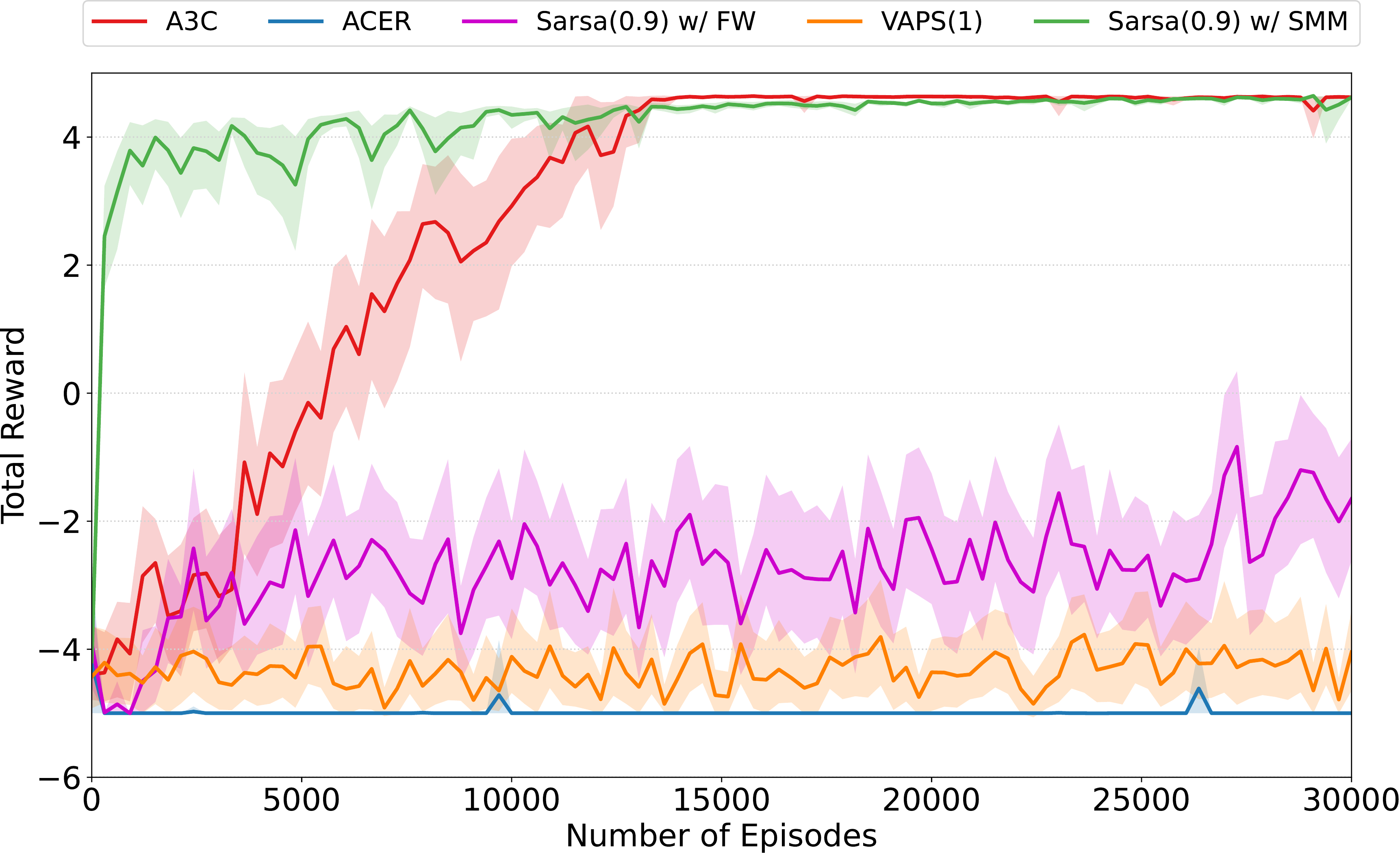}
        \label{fig:total_reward_meuleau_maze_v2}
	}
	\subfloat[]{
	    \centering
		\includegraphics[scale=0.2]{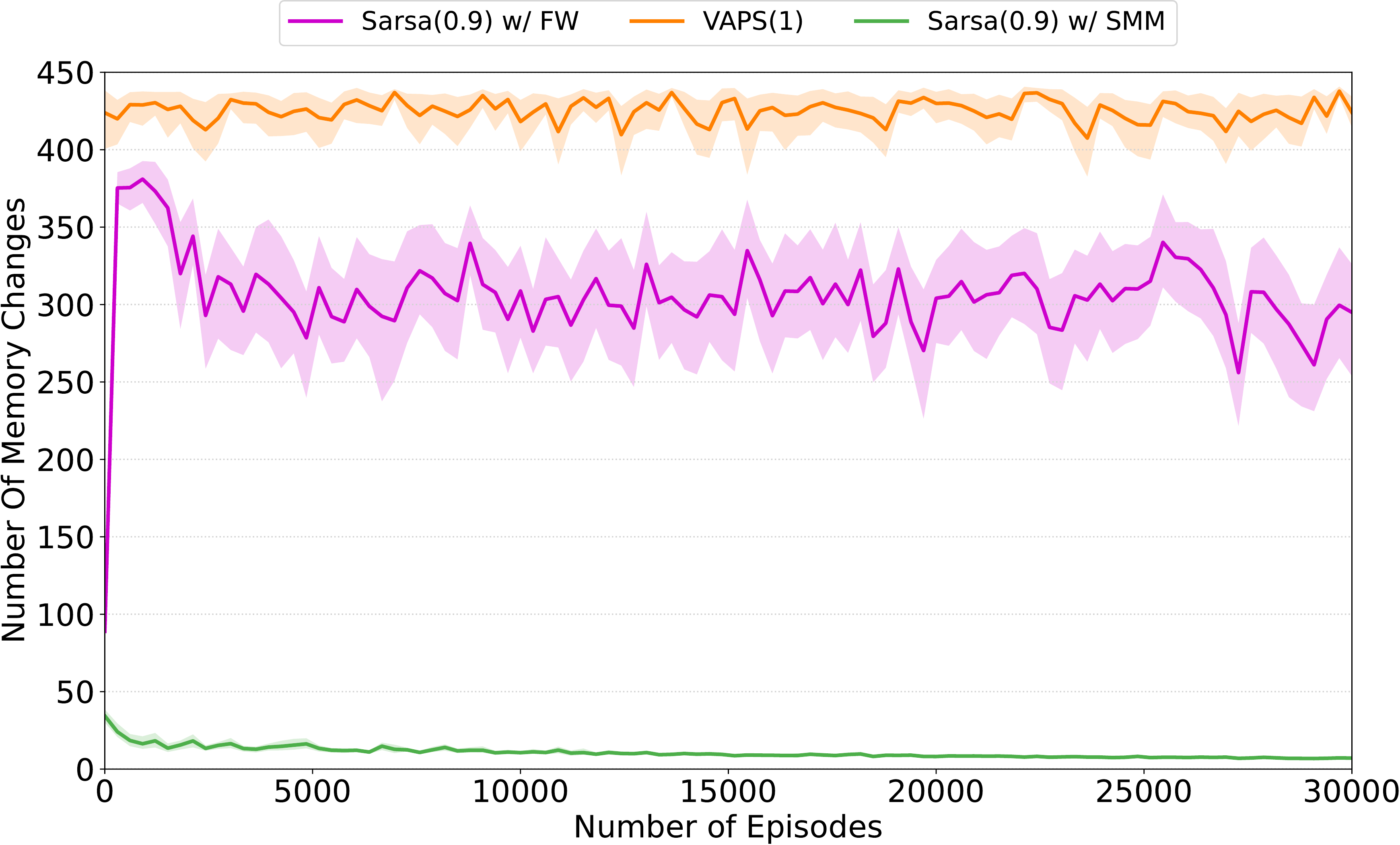}
		\label{fig:memory_changes_meuleau_maze_v2}
	}
	\caption{(a) Total reward collected throughout an episode, (b) number of changes on external memory during an episode in \texttt{Modified Meuleau's Maze} problem. VAPS(1) algorithm has a 3 bit memory where Sarsa($\lambda$) with FW and Sarsa($\lambda$) with SMM has a memory with a capacity of 3 and 1 observations, respectively. The lines represent the average of 50 experiments and shaded areas are the $95\%$ bootstrapped confidence intervals.}
	\label{fig:result_meuleau_maze_v2}
\end{figure}

In this problem, we experimented on Sarsa($\lambda$) with FW and VAPS(1) with different memory capacities, yet they were unable to get a good learning performance. Figure \ref{fig:total_reward_meuleau_maze_v2} shows the best achieved performance among several configurations for these methods. In the figure, the effect of SMM supported by intrinsic motivation is clear. Sarsa($\lambda$) with SMM, displays a rapid learning and converges to a near optimal performance in this stochastic environment whereas Sarsa($\lambda$) with FW and VAPS(1) fail to do so, even with a bigger memory capacity. While ACER is unable to reach to the goal state, A3C also learns the best policy but its convergence is slower compared to SMM. Faster convergence of SMM is attained by keeping the observations at the corners and T-junctions, which are relatively uncommon, in the memory. For example, while being in a location inside a vertical corridor is ambiguous, keeping a bottom-left corner observation in memory allows the agent to go north, then having a T-junction observation with north direction blocked allows it to go south. This way, the agent learns to update its memory repeatedly so that it reaches the goal. Keeping a memory based on a window is not beneficial under this setting, since the temporal distance to key observations varies on different states of the problem. The performance of VAPS(1) algorithm shows that it is restricted to deterministic domains with sparse rewards. While the other external memory based approaches constantly update the memory, SMM achieves a good learning performance with minimum memory changes, shown in Figure \ref{fig:memory_changes_meuleau_maze_v2}. 

\subsection{\texttt{Modified Tree Maze}}
\begin{figure}[h!]
    \centering
    \includegraphics[scale=0.3]{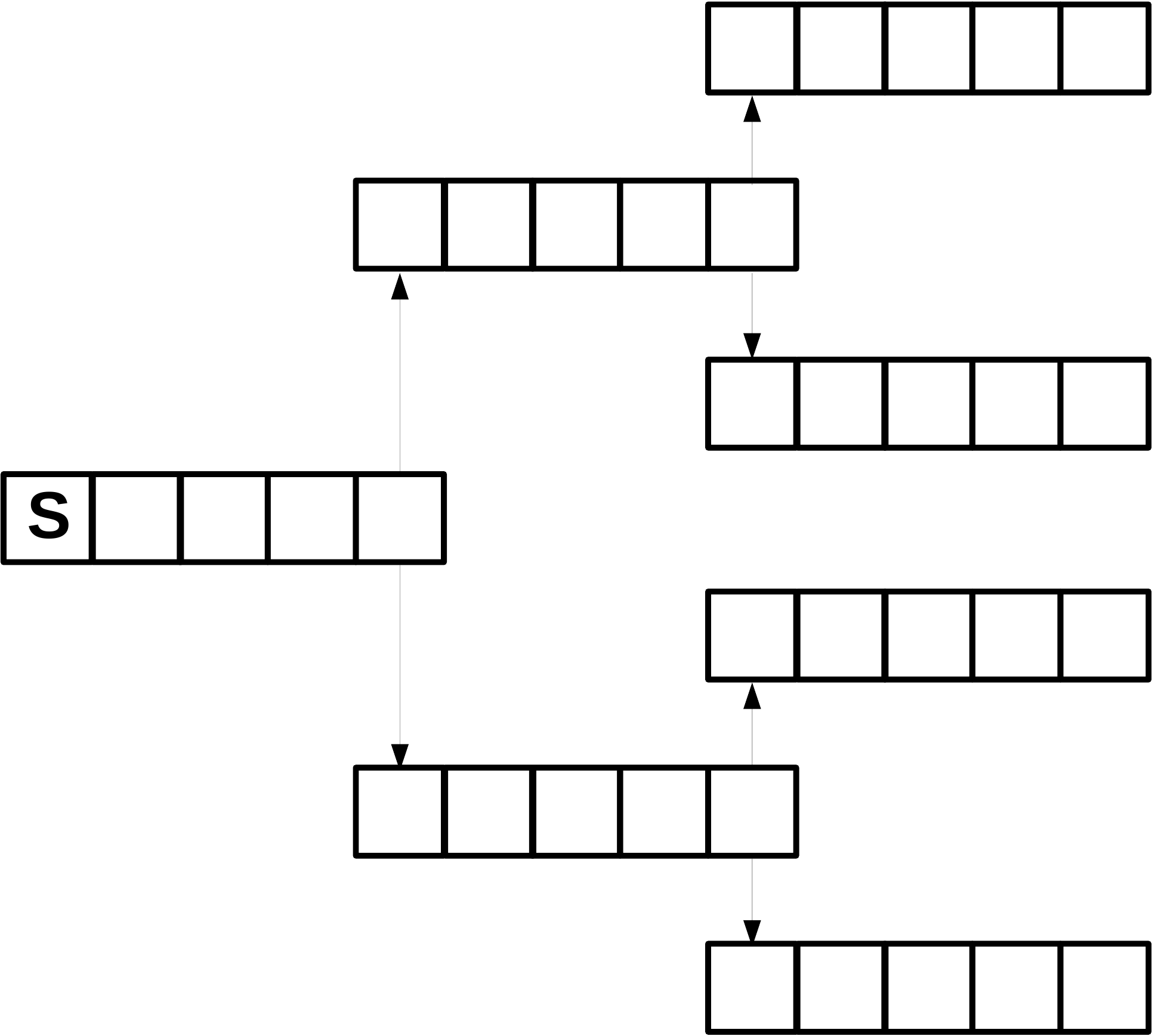}
    \caption{Sketch of \texttt{Modified Tree Maze} where the agent starts from the state with $S$ and aims to reach to a leaf which is determined at the beginning of each episode. The agent observes the turns it needs to take only at the first two steps of the episode.}
    \label{fig:tree_maze}
\end{figure}

\texttt{Modified Tree Maze}, shown in Figure \ref{fig:tree_maze}, is another partial observable problem that contains long-term dependency \citep{Steckelmacher2018}. The agent starts from the location $S$ and needs to reach to a state at the end of a leaf. It has three deterministic actions to go \textit{north}, \textit{east}, \textit{south}. The goal state to reach is determined at the beginning of each episode, making the problem more complex. An episode ends with a reward of 10 when the agent reaches to the goal state and -0.1 for reaching to other leaf states. All regular steps of the agent are also rewarded by -0.1. An observation is limited to the location in the corridor (represented as left-most, middle and right-most), the number of taken turns and an additional bit to represent the required turns to take in each T-junction, which are only available in the first two steps of the episode. In this way, the agent has to keep these first observations in memory until they are useful, in order to reach to the desired goal state. As a result, the environment has $|S| = 140, |A| = 3, |\Omega| = 14$.

In this problem, we allowed VAPS(1) to have a 2 bits of memory while FM and SMM had a memory capacity of 2 observations so that all external memory based methods have enough capacity to learn the task.

\begin{figure}[t!]
	\centering
	\subfloat[]{
	    \centering
		\includegraphics[scale=0.2]{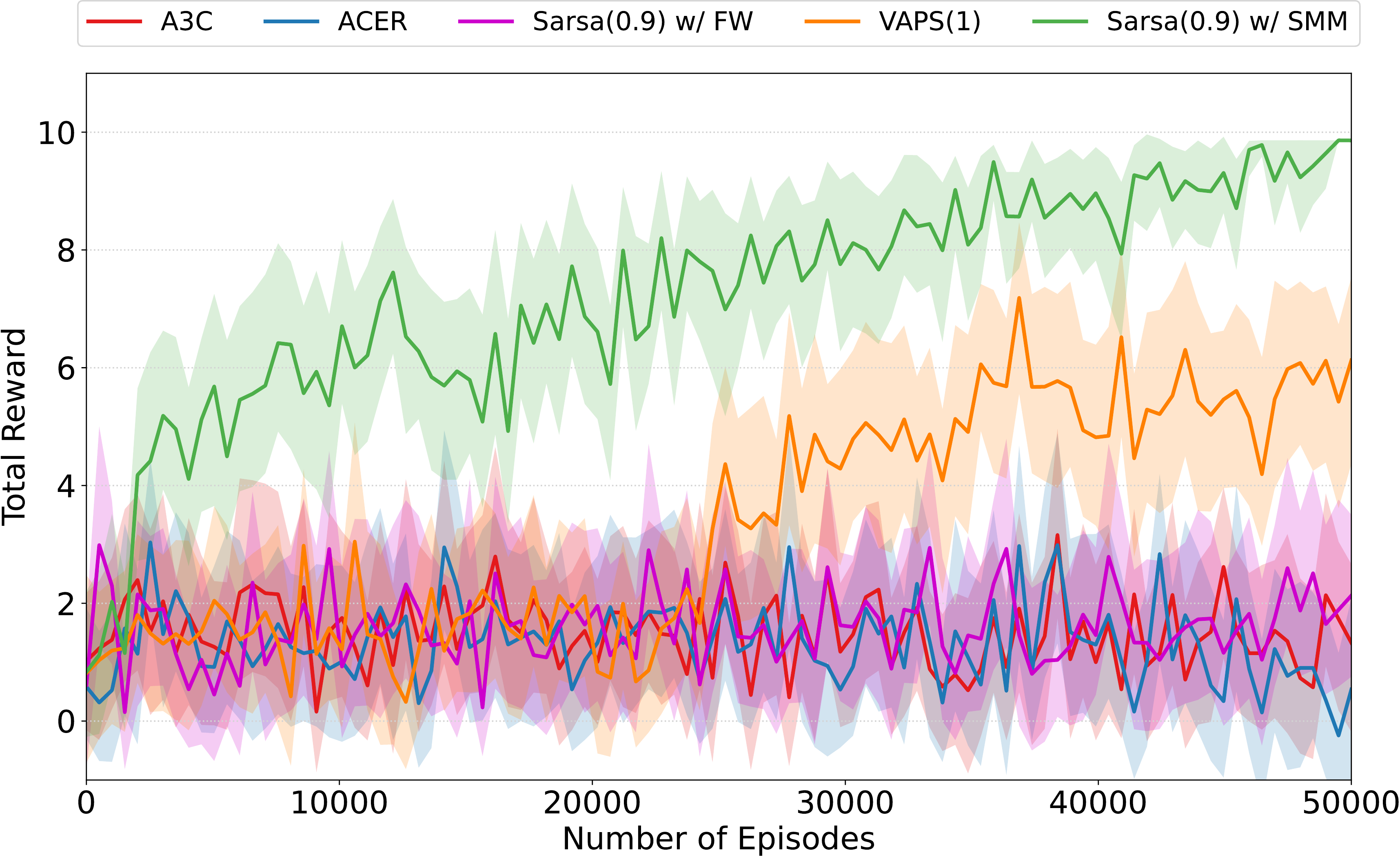}
        \label{fig:total_reward_tree_maze_v2}
	} 
	\subfloat[]{
	    \centering
		\includegraphics[scale=0.2]{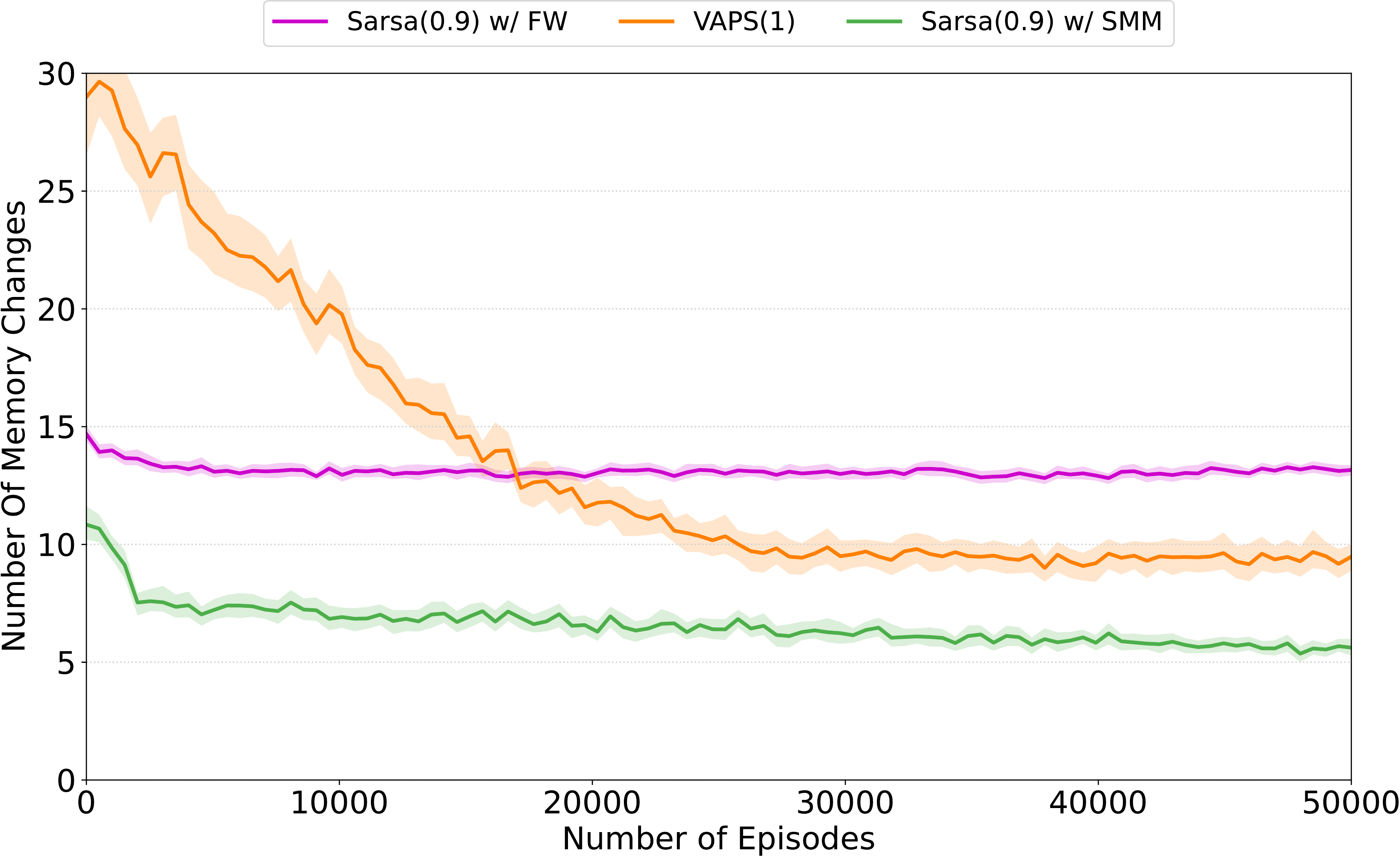}
		\label{fig:memory_changes_tree_maze_v2}
	}
	\caption{(a) Total reward collected throughout an episode, (b) number of changes on external memory during an episode in \texttt{Modified Tree Maze} problem. VAPS(1) algorithm has a 2 bit memory where both Sarsa($\lambda$) with FW and Sarsa($\lambda$) with SMM has a memory with a capacity of 2 observations. The lines represent the average of 50 experiments and shaded areas are the $95\%$ bootstrapped confidence intervals.}
	\label{fig:result_tree_maze_v2}
\end{figure}

As shown in Figure \ref{fig:total_reward_tree_maze_v2}, only VAPS(1) and Sarsa($\lambda$) with SMM learns in this problem while the others show none but random achievements. Sarsa($\lambda$) with SMM outperforms all by reaching to a near optimal performance with a total reward of 9.86 in the last episodes. The methods using LSTM, A3C and ACER, cannot devise a good policy, supporting the argument that they require specific configurations for a problem to learn it. Similar to the other results, the number of changes that SMM applies to the memory converges to a minimum value, as Figure \ref{fig:memory_changes_tree_maze_v2} shows. Note that SMM does not fix the content of the memory at the beginning of an episode in this setting, but updates it when the corresponding observation is no longer required to solve the task. For example, the first observation in the episode contains the direction that the agent needs to take in the first T-junction. SMM removes this observation from the memory by storing a new one, when the agent takes that turn. This shows that SMM learns a fitting behaviour for the task.

\begin{figure}[h!]
	\centering
	\subfloat[]{
	    \centering
		\includegraphics[scale=0.2]{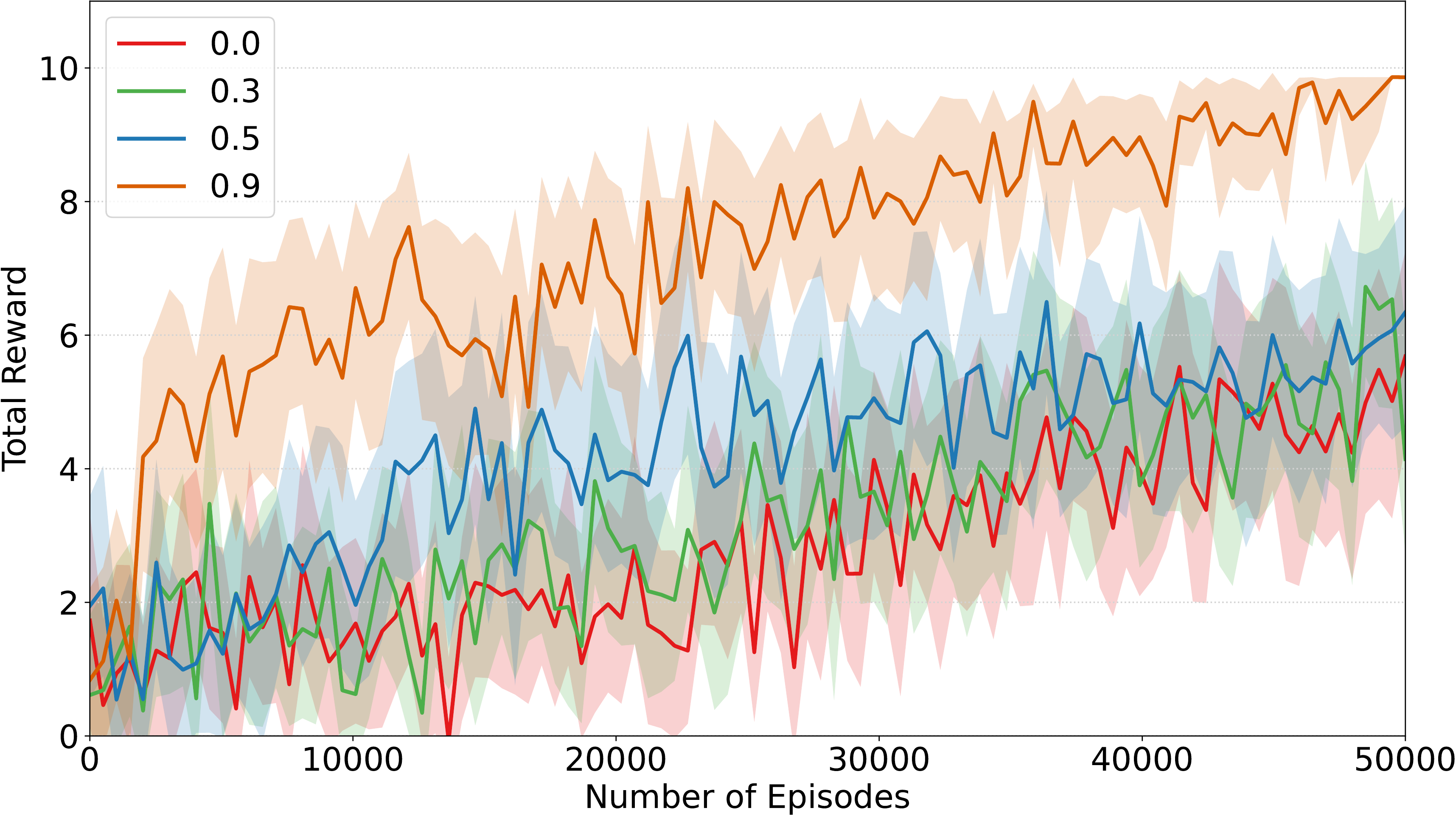}
        \label{fig:total_reward_tree_maze_v2_lambda}
	}
	\subfloat[]{
	    \centering
		\includegraphics[scale=0.2]{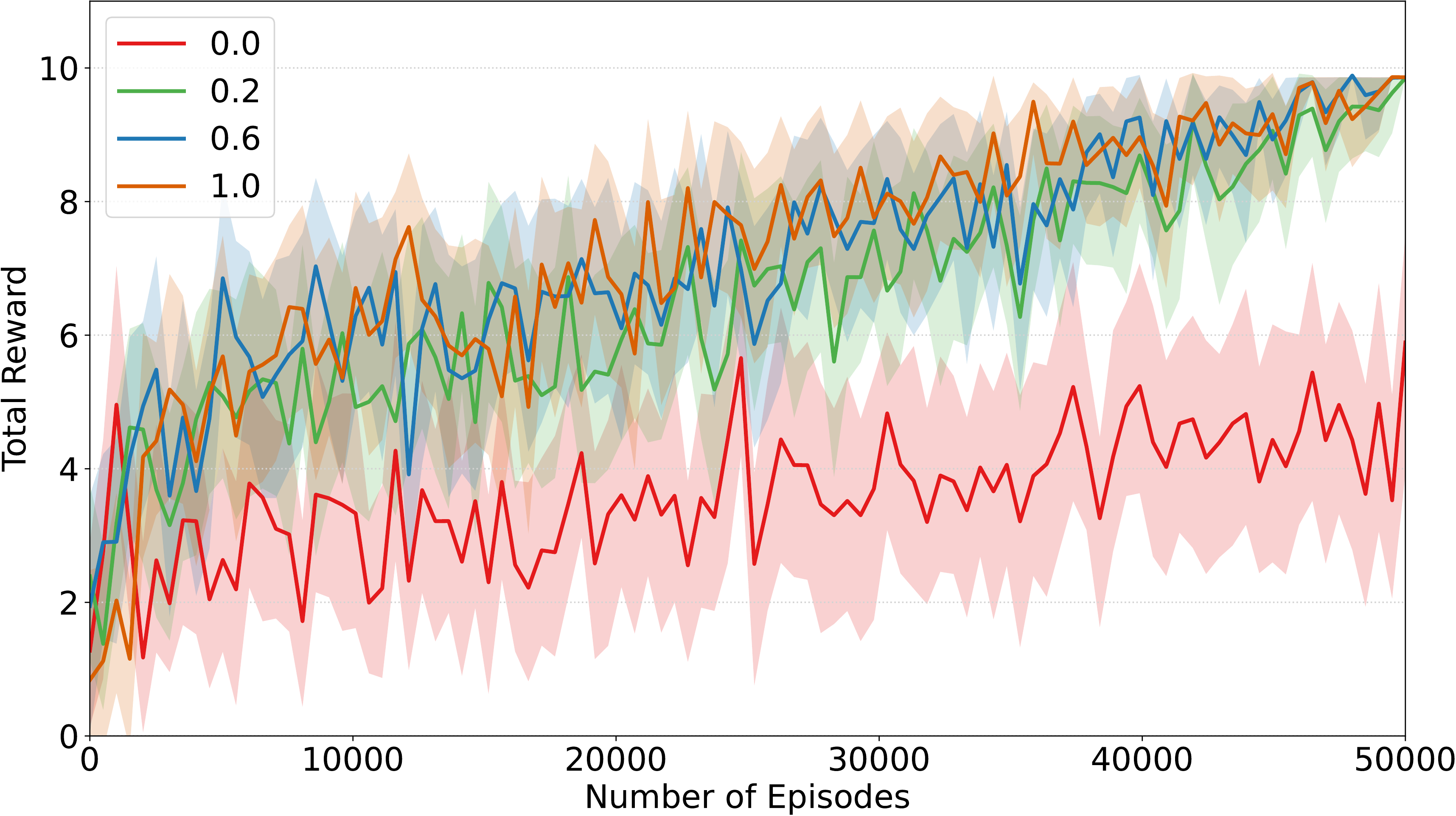}
		\label{fig:total_reward_tree_maze_v2_beta}
	}
	\caption{(a) Total reward collected by Sarsa($\lambda$) with SMM on different $\lambda$ values, (b) total reward collected by Sarsa(0.9) with SMM on different $\beta$ values in \texttt{Modified Tree Maze} problem. Sarsa($\lambda$) with SMM has a memory with a capacity of 2 observations. The lines represent the average of 50 experiments and shaded areas are the $95\%$ bootstrapped confidence intervals.}
	\label{fig:result_tree_maze_v2_lambda_beta}
\end{figure}

We further tested Sarsa($\lambda$) with SMM on several values of its parameters in \texttt{Modified Tree Maze}. To avoid overcrowding in the figures, we kept this set of values small. Figure \ref{fig:total_reward_tree_maze_v2_lambda} shows the effect of various $\lambda$ values on the learning performance. As expected, lower $\lambda$ translates to a slower convergence to a good policy since higher $\lambda$ allows Sarsa($\lambda$) algorithm to propagate the TD-error to a larger set of eligible estimated state - composed action pairs. In Figure \ref{fig:total_reward_tree_maze_v2_beta}, the effect of $\beta$, therefore the amplitude of intrinsic rewards, on learning can be seen. While learning with $\beta = 0$ is slow, even a small value as $\beta = 0.2$ improves the learning speed of SMM. Here, even though the intrinsic reward is small, it can modify the policy so that the agent is motivated to keep rare observations in memory, leading to a greater learning performance.

\section{Conclusion}
This study proposes a main layout where a reinforcement learning agent learns how to manage its own memory with an intrinsic motivation to keep useful observations in it. The proposed method, self memory management, defines the form of the memory, the actions changing the memory and the intrinsic reward to be provided to the learning updates. It can be coupled with many reinforcement learning algorithms and is shown to work in partially observable environments that have long-term dependency. In the experiments, we compare our method with the original study that proposes the idea to allow the agent to control the memory and show that with the help of intrinsic motivation, it outperforms several memory based approaches in various POMDPs, with minimal changes on the memory.

There are ways to do further research on this paper, which acts as a preliminary study. First, the set of memory actions $\mathcal{A}$ can be extended to give more control to the agent and the capacity bounding parameter $c$ can be disregarded, giving the full control of the memory size to the agent. Second, the definition of intrinsic reward can be altered as some problems may require additional heuristics on what to keep in the memory, besides the rare observations. Finally, the more memory based approaches in the literature can be included in the experiment set.

\section*{Acknowledgements}
This work is supported by the Scientific and Technological Research Council of Turkey under Grant No. 120E427. Authors would also like to thank H\"{u}seyin Ayd{\i}n, Erkin \c{C}ilden and Faruk Polat for their support.

\bibliographystyle{abbrvnat}
\bibliography{references}

\end{document}